\documentclass[runningheads]{llncs}
\usepackage{graphicx}
\usepackage{amsmath,amssymb} 
\usepackage{array}
\usepackage{arydshln}
\newcolumntype{P}[1]{>{\centering\arraybackslash}p{#1}}
\usepackage{algorithm}
\usepackage{algpseudocode}
\usepackage{color}

\begin{document}
\pagestyle{headings}
\mainmatter


\title{Explaining Deep Neural Networks for Point Clouds using Gradient-based Visualisations} 
\titlerunning{Explaining DNNs for Point Clouds using Gradient-based Visualisations}
\authorrunning{Jawad Tayyub, Muhammad Sarmad, Nicolas Schönborn}

\author{$^1$Jawad Tayyub$^*$, $^2$Muhammad Sarmad$^*$, $^1$Nicolas Schönborn$^*$ \\ \small{$^*$equal contribution}}
\institute{$^1$Endress + Hauser, Maulburg, Germany \\
$^2$Norwegian University of Science and Technology, Trondheim, Norway \\
}


\maketitle

\vspace*{-0.3cm}%

\begin{abstract}
   Explaining decisions made by deep neural networks is a rapidly advancing research topic. In recent years, several approaches have attempted to provide visual explanations of decisions made by neural networks designed for structured 2D image input data. In this paper, we propose a novel approach to generate coarse visual explanations of networks designed to classify unstructured 3D data, namely point clouds. Our method uses gradients flowing back to the final feature map layers and maps these values as contributions of the corresponding points in the input point cloud. Due to dimensionality disagreement and lack of spatial consistency between input points and final feature maps, our approach combines gradients with points dropping to compute explanations of different parts of the point cloud iteratively. The generality of our approach is tested on various point cloud classification networks, including 'single object' networks PointNet, PointNet++, DGCNN, and a 'scene' network VoteNet. Our method generates symmetric explanation maps that highlight important regions and provide insight into the decision-making process of network architectures. We perform an exhaustive evaluation of trust and interpretability of our explanation method against comparative approaches using quantitative, quantitative and human studies. All our code is implemented in PyTorch and will be made publicly available.
\end{abstract}

\section{Introduction}
\label{sec:intro}


The black-box nature of deep neural networks is a major hindrance in their utilization and wide-acceptance in real-world and safety-critical scenarios. This trust deficit can be mitigated by interpreting the reasoning behind a network's behaviour. Researchers have made significant progress in proposing explanation methods \cite{gradients,CAM,gradCAM,lrp} for demystifying CNN networks for image processing. However, interpretation of deep networks designed for 3D data, namely point clouds \cite{pointnet,pointnet++,dgcnn,votenet}, remain an understudied area. Point clouds are an unstructured representation as opposed to regular grids such as images or voxel grids. Therefore, direct application of existing interpretation methods for image-based deep networks is not suitable for point cloud deep networks.

\begin{figure}
  \centering
   \includegraphics[width=1.0\linewidth,trim={0 0 0 0},clip]{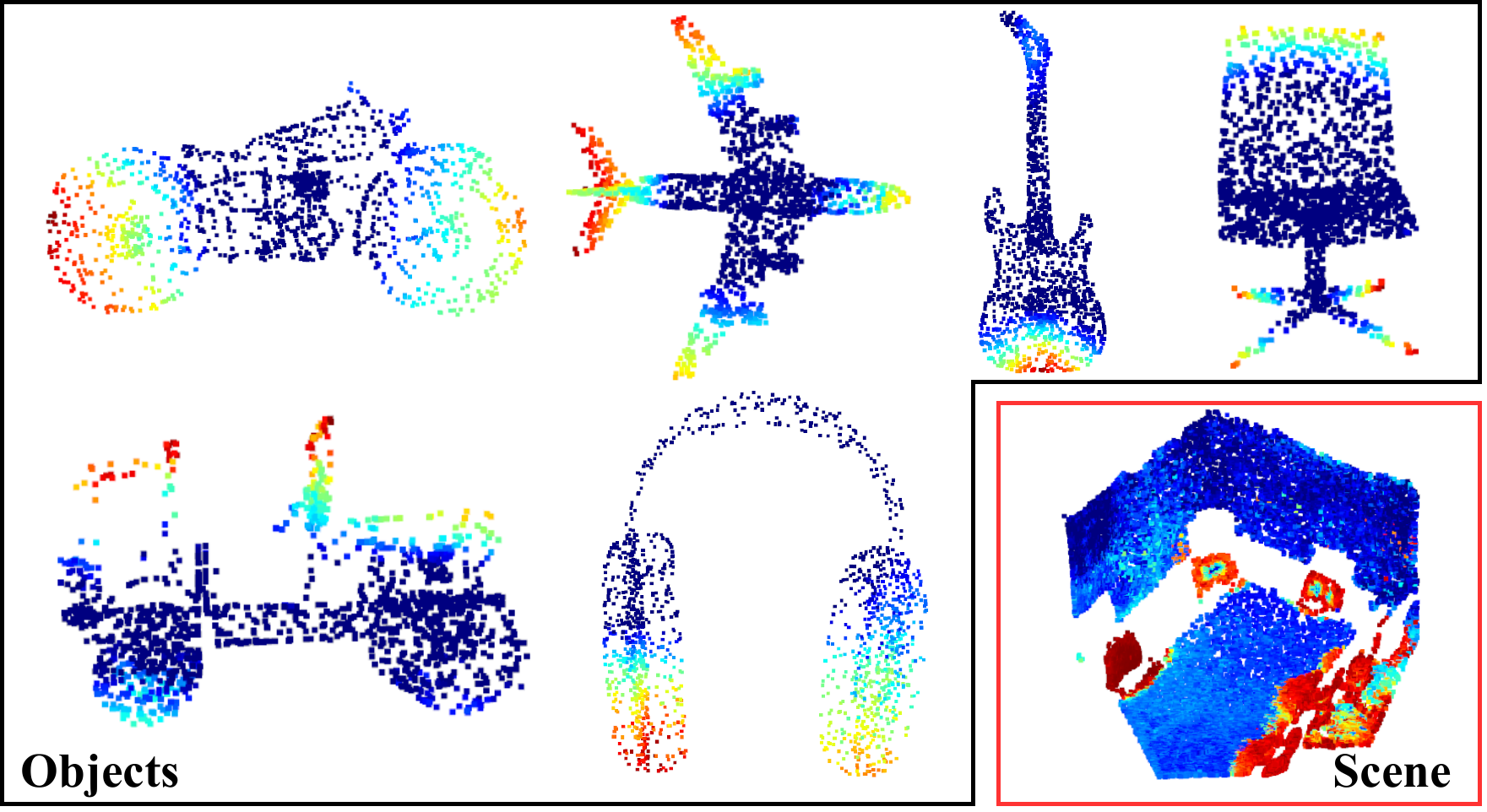}
   \caption{\small{\textbf{Point Cloud Heatmaps} Our proposed approaches highlight the salient regions in single object and scene point clouds that are critical for the decision-making of a point cloud processing network.}}
\label{fig:teaser}
\end{figure}


Firstly, typical CNNs progressively apply convolution and pooling operations to images resulting in low-resolution final feature maps while preserving spatial consistency. Therefore, explanation values generated through logging gradients back to final feature layers of CNN's \cite{gradCAM,CAM} can be mapped to the input image through bi-linear scaling. For unstructured point cloud data, such spatial consistency cannot be guaranteed, and therefore mapping logged gradients back to the input points is non-trivial. Second, the correspondence between each input point and the final feature layer neurons cannot be asserted in point cloud networks. Furthermore, image based explanation techniques such as gradients \cite{gradients} or deconvolutions \cite{zeiler2014visualizing} operate directly at the input pixel-level resulting in fine and grainy explanations, making them difficult for humans to interpret. 

In this work, we address these challenges and propose accumulated piecewise explanations (APE) which is a general explanation approach applicable to a wide variety of deep networks for point clouds. APE computes a point cloud heatmap which highlights each input point's contribution towards the network's decision. The generated heatmaps are interpretable and provide visual insight into the network's behavior. Our method logs gradients (computed for a preset target class) from each neuron in the final feature extraction layer of the network. These are then mapped to the input point cloud to generate the point cloud heatmap as seen in Fig.\ref{fig:teaser}. Gradients are computed w.r.t. the final feature maps of a point cloud processing network architecture. These values can be used to demystify a network's inner workings. Since the feature extraction layers of a network reduce resolution through pooling operations, a direct mapping only reveals an explanation for a small segment of the point cloud. To resolve this, we propose to iteratively explain segments of the point clouds by dropping explained points from the previous iteration, allowing for explanation values to be computed for a different segment. Heatmaps gathered from each iteration are concatenated to generate a complete heatmap. Finally, this heatmap is refined through a second iterative process that recomputes heatmaps while dropping the lowest relevant points from the previous iteration. A weighted maximum over heatmaps from all iterations yields a high-fidelity point cloud heatmap. A key feature of our approach is that the generated heatmaps highlight semantic segments of a 3D shape regardless of the network architecture and therefore exhibit human friendly visual representation. We refer to this property as 'human interpretability'. The heatmaps generated by our method do not just look aesthetically pleasing, but also correctly highlight critical points. This is verified quantitatively in experiments. Our contributions are summarised below:

\begin{itemize}

\item We propose a general algorithm to explain 3D point cloud deep networks by generating human interpretable heatmaps.

\item We extensively evaluate our explanation heatmaps on various point cloud classification architectures, namely PointNet, PointNet++, DGCNN, and VoteNet. Deteriorating performance of networks is shown by dropping high relevance points identified by our approach. We also evaluate our method against a existing work and demonstrate SOTA performance.

\item We demonstrate that our approach outputs higher fidelity heatmaps than existing explanation methods for point clouds and images. Moreover, our strategy highlights biases and failure modes of point cloud networks and provides in-depth insights.

\end{itemize}

\section{Related Work}
\label{sec:relatedworks}

\paragraph{Deep Learning on Point Cloud}
Processing a 3D point cloud directly via deep networks has recently gained attention. Much work in this area has surfaced attempting to solve a wide array of vision problems for point clouds \cite{pointnet,pointnet++,dgcnn,rlgan,panos,splat,yolo}. Our work aims to analyse these opaque models and generate visual explanations for them. Seminal work on point clouds classification includes \cite{pointnet,pointnet++,votenet,dgcnn}. We demonstrate that these approaches benefit from our explainability method, providing clear insight into the network's decision-making process.

\paragraph{Explainability Methods for CNNs}
The availability of large datasets \cite{imagenetData,coco,modelnet} and discovery of CNNs have provided breakthroughs on various challenges in the vision community  \cite{resnet,imagenetCls,cnn2}. Efforts to visualize CNNs date back to their discovery, and many methods have been proposed in literature \cite{gradCAM,CAM,objectlocfree,gradients,ziwen2019visualizing,pinherio,lrp,rise}. Pope et al. \cite{graphGrad} have extended CNN explainability methods to graph CNNs. Selvaraju et al. \cite{gradCAM} proposed Grad-CAM which generates gradient-based visual explanations. Grad-CAM proposes to log gradients at intermediate layers rather than input layers to generate a heatmap that reflects each activation's importance. This heatmap is overlaid onto the input image by simple bilinear scaling to represent contributions of pixels in the image for a given task, e.g., classification. Since later feature layers capture higher-level semantic features, observing these gradients allows for computing coarse explanation values that are human-interpretable. However, application of image-based explanation approaches for point clouds is non-trivial due to their unstructured nature. 


\paragraph{Explainable Methods for Point Cloud Based Deep Networks}
There is little work on visualizing deep networks that process 3D data such as point clouds. Zhang et al. \cite{explain_pointnet} have visualized PointNet; however, they utilize class attentive features and modify PointNet's architecture. Our technique does not require any modification in the architecture. Qiu et al. \cite{qiu2019geometric} visualize the layers of their proposed point cloud processing pipeline, but they do not employ any gradient-based visualizations.  Zheng et al. \cite{pcSaliencyMap} propose a differentiable point shifting process that simulates point dropping and computes contribution scores to input points according to the loss value. This approach identifies highly accurate contributions of individual points; however, it compromises global interpretability. Computed heatmaps resemble fine details similar to gradients \cite{gradients} or LRP \cite{lrp} methods in images.  In contrast, our method offers semantic visualisations which provide a global overview of the network's focal points. We argue that a heatmap that highlights point cloud's segments that correspond to semantic parts of a shape, e.g., the legs of a chair, builds higher trust than fine-grained explanations.


\section{Preliminaries}

\begin{figure*}
\begin{center}
   \includegraphics[width=1.03\textwidth]{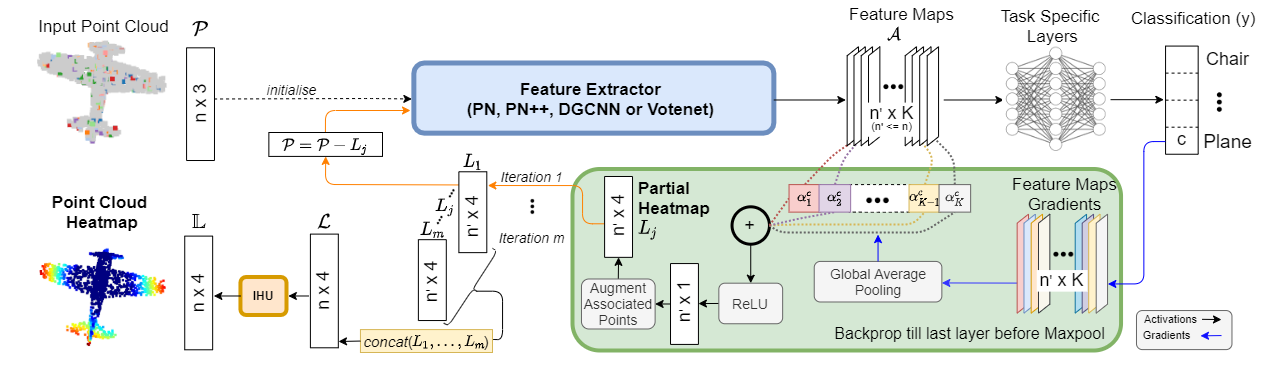}
\end{center}
   \caption{\small{{\bf Approach Overview.} Our approach maps gradients logged at the feature maps layer to the point cloud for a target class e.g. `Airplane'. Gradients are computed w.r.t to the feature maps $\mathcal{A}$. $L_j$ denotes the computed partial heatmap with elements having the form $(P_i,l_i)$ where each point coordinate is augmented with an explanation value. This is a weighted sum of the feature maps followed by ReLU. This partial heatmap only explains $n'$ points of the point cloud since there are $n'$ features per feature map. Explained points in $L_j$ are removed from $P$ iteratively to produce $m$ partial heatmaps. A concatenation of all these partial heatmaps result in a complete \textit{initial} heatmap $\mathcal{L}$. An IHU step, detailed in text, refines the initial heatmap to produce the final point cloud heatmap $\mathbb{L}$.}}
\label{fig:point-grad-cam}
\end{figure*}

A point cloud is defined as a set ${\mathcal{P} = \{P_1,...,P_n\}}$ of $n$ points where each point is denoted by its 3D coordinates $(x, y, z)$. A deep neural network $f: \mathcal{X} \rightarrow C$ is any trained classifier which maps an input point cloud $\mathcal{P} \in \mathcal{X}$ to a class $c \in C$. Then, given a target class $c$, our goal is to find an explanation heatmap ${\mathbb{L} = \{(P_1,\ell_1), ..., (P_n,\ell_n)\}}$ where each $\ell_i \in [0,1]$ represents the \textit{contribution} of corresponding point $P_i$ towards the network's decision. $\mathbb{L}$ denotes the point cloud heatmap of the input point cloud $\mathcal{P}$.

Point cloud classification networks are categorised into fixed and variable networks. As illustrated in Fig. \ref{fig:point-grad-cam}, an input point cloud ${\mathcal{P} \in \mathbb{R}^{n\times3}}$ of $n$ points transforms into $K$ feature maps $\mathcal{A} \in \mathbb{R}^{n'\times K}$ each of length $n'$. Fixed networks, such as PointNet \cite{pointnet} or DGCNN \cite{dgcnn}, preserve the dimensionality of $\mathcal{P}$ and subsequent feature maps. In these networks, convolutions are applied to each point resulting in a feature map per point hence satisfying $n = n'$. In variable networks, such as PointNet++ \cite{pointnet++} and VoteNet \cite{votenet}, feature maps in subsequent layers reduce or accumulate input points by various methods e.g sampling and grouping, clustering, etc. In these networks, $n'$ is often less than $n$. Most other models either fall into the first or the second category depending on their design.

We also define a \textit{point dropping} operation as the removal  of points from a point cloud $\mathcal{P}$. For consistency, our point dropping mechanism is similarly to \cite{pcSaliencyMap} whereby point coordinates are shifted to the spherical core (centre) of the point cloud $\mathcal{P}$ neutralising their effect to a high degree.

\section{Accumulated Piecewise Explanations (APE)}
\label{sec:method}

In this section, we present our approach called accumulated peiecewise explanations for generating highly interpretable point cloud heatmaps $\mathbb{L}$, Fig. \ref{fig:point-grad-cam}. An input point cloud $\mathcal{P}$ is first classified using a network $f$, such as PointNet, PointNet++ etc. The final feature maps $\mathcal{A}$, before task-specific fully connected layers, capture high-level semantics which are used for generating coarse conceptual explanations. Therefore, we compute gradients with respect to the final feature maps $\mathcal{A}$. These feature map gradients can be used to gain insight into the network's decision. The gradients are then globally average pooled (GAP) per feature map to create a weighting $\alpha$ which reflects the contribution of neurons in each feature map. A \textit{partial heatmap} $L_j$ is then constructed by taking a weighted sum of feature maps $\mathcal{A}$ representing the aggregated contribution of corresponding input points $n'$. However, since feature extraction layers are of a lower resolution from the input point cloud, partial heatmap $L_j$ at this stage only reveals contributions of a subset of the point cloud $\mathcal{P}$. To generate contributions of the other segments of $\mathcal{P}$, the subset of points explained by $L_j$ are dropped from $\mathcal{P}$ iteratively where $j$ denotes the iteration. This allows for explanation values to be computed for a different segment. All partial heatmaps $(L_1,...,L_m)$ are then concatenated to form an initial point cloud heatmap $\mathcal{L}$. A second iterative process called iterative heatmap update (IHU) refines this output by iteratively dropping lowest relevance points to generate a high quality final point cloud heatmap $\mathbb{L}$. This process is further detailed in subsequent section. 

Algorithm \ref{algo:APE} presents a formal description of our APE method. This algorithm comprises of two nested loops. The inner loop computes partial heatmaps $L_j$ from feature maps $A$ whereas the outer loop refines these heatmaps $\mathcal{L}$ by dropping lowest relevance points $n_L$ iteratively recomputing $\mathcal{L}$ in each iteration. This process iterates $\lambda$ times and heatmaps gathered from different iterations are combined by a weighted maximum to produce the final point cloud heatmap $\mathbb{L}$. Next, we explain the partial heatmap $L_j$ compution followed by the complete point cloud heatmap $\mathbb{L}$ in higher detail.

\begin{algorithm}
\caption{Accumulated Piecewise Explanations (APE)}\label{algo:APE}

\begin{algorithmic}
\State \textbf{Require:} $\lambda$: number of IHU iterations. $c$: target class. 
\end{algorithmic}

\begin{algorithmic}
\State  \textbf{Input:} Point cloud $\mathcal{P}$, Point Cloud classifier $f$.
\end{algorithmic}

\begin{algorithmic}
\State  \textbf{Output:} Point Cloud heatmap $\mathbb{L}$.
\end{algorithmic}

\begin{algorithmic}[1]
 
\For{$i = 1....\lambda$}
\State$j = 0$
\While{$\mathcal{P}$ is not empty}
\State $y^c=f(\mathcal{P})$
\State $\alpha^c_k = \frac{1}{n'} \sum_{n'} \frac{\delta y^c}{\delta A^k_{n'}}$
\State $L_j = ReLU(\sum_k \alpha^c_k A^k)$
\State $\mathcal{P} = \mathcal{P} - L_j$ (Drop explained points)
\State Increment $j$

\EndWhile
\State $\mathcal{L}_i = concatenate(L_1, L_2, ..., L_m)$
\State Drop $n_L$ lowest contribution points from  $\mathcal{P}$. 
\EndFor
\State $\mathbb{L} = \max_{i=1,..,\lambda}w_il^i_j, \forall j=1,..,m$  where $l^i \in L^A_i$
\end{algorithmic}
\label{algo:algo1}
\end{algorithm}

\subsection{Partial Heatmap $L_j$}

To compute the partial heatmap $L_j$, first gradient values corresponding to each of the neurons in the final feature map $\mathcal{A}$ are computed. Algorithm \ref{algo:APE} details this step in lines 4-6. Input point cloud $\mathcal{P}$ is classified to produce $y^c$ which denote the classification score of a given target class $c$ (line 4). Gradients are then computed w.r.t to the final feature maps $\mathcal{A} = \{A^1,...,A^K\}$. Each feature map $A^k \in \mathbb{R}^{n'}$ is of length $n'$. The calculated gradients are then globally average pooled (GAP) per $A^k$ to generate a contribution weight $\alpha^c_k$ for the $k^{th}$ feature map (line 5). Point cloud $\mathcal{P}$ has $n$ points whilst $\frac{1}{n'} \sum_{n'}$ is global average pooling and $\frac{\delta y^c}{\delta A^k_{n'}}$ are the gradients. This results in a weight vector where each weight $\alpha^c_k$ gives a relevance weighting of the neurons in the $k^{th}$ feature map. The \textit{partial heatmap} $L_j$ is then computed by taking the positively contributing gradients only through utilising $ReLU$ (line 6) normalized in the range $[0,1]$. $L_j$ computed at this stage reflects the contributions of $n'$ neurons in the final feature maps and not the input point cloud $\mathcal{P}$. 

\subsection{Point Cloud Heatmap}


Depending on the network architecture (fixed or variable), $L_j$ may be of lower dimensionality than $\mathcal{P}$, i.e. $n'\leq n$. We utilise an iterative mechanism whereby explained points are dropped in each iteration, and a new partial heatmap is computed for a different segment of the input point cloud. This is repeated until explanation values are computed for the complete point cloud.
This method provides an exact contribution estimate for each point compared to trivial interpolation. Due to tracked associations from $L_j$ to $\mathcal{P}$, the subset of points explained by $L_j$ are identified and dropped from $\mathcal{P}$ (line 7). The resulting point cloud is again passed as input to the network in the next iteration to generate a new partial heatmap $L_{j+1}$ explaining a different subset of points. This process is repeated $(j = 1,...,m)$ until all points in the input point cloud have been explained. A concatenation of $L_j$ from all iterations produces an initial point cloud heatmap $\mathcal{L}$ (line 10). Note that we concatenate the raw value of the partial heatmap and then normalise instead of normalising and concatenating. This procedure is valid since it preserves the relative importance of a particular point w.r.t others.

To create the final point cloud heatmap $\mathbb{L}$, the IHU step is introduced. This step computes initial point cloud heatmaps $\mathcal{L}_i$ iteratively, whereby in each $i^{th}$ iteration, lowest relevance value points $n_L$ are dropped from $\mathcal{P}$ (line 11) where $n_L$ is a empirically set hyperparameter. The resulting point cloud is reclassified to compute a new initial point cloud heatmap $\mathcal{L}_{i+1}$. After $\lambda$ iterations, all points have been dropped from $\mathcal{P}$ and $\lambda$ feature heatmaps $(L_1,...,L_\lambda)$ have been computed. The final point cloud heatmap $\mathbb{L}$ is then computed by merging the different initial heatmaps (line 13), where $l^i$ are components of $\mathcal{L}_i$ and weights $w$ are hyperparameters which are empirically set. Fig. \ref{fig:visualpointdrop} presents a visual illustration of this process. Removing the least significant points in each iteration allows for a better explanation of the remaining points. This enhances explanations by highlighting significant salient areas of the point cloud, which are suppressed by low-contributing points. Explanations over these suppressed areas are revealed by removing low-contributing points iteratively. Also, by taking the max explanation value of points over all iterations, the discovery of most liberal explanation is ensured.

\paragraph{Fixed and Variable Network Architectures}

The proposed algorithm is applicable for both fixes and variable network architectures. Fixed networks preserve the dimensionality of input point cloud $\mathcal{P}$ to the feature maps $\mathcal{A}$, a simple bijective mapping of $\mathcal{A}$ to $\mathcal{P}$ is sufficient to create a final point cloud heatmap since $n'=n$. Therefore the inner loop only has one iteration in these cases, namely PointNet and DGCNN. Note that the partial heatmap module can be applied to any layer before task-specific layers to generate the final point cloud heatmap. Variable networks produce lower-dimensional feature maps. This property is similar to image domain CNNs where, through convolution and pooling operations, computing an explanation heatmap at any intermediate layer results in a lower resolution heatmap than the input image. Similarly, parital heatmaps computation on intermediate feature maps of variable networks for point clouds produce a sparse heatmap since $n' \leq n$. In the case of images, since CNN preserves spatial consistency, a simple bi-linear scaling of partial heatmaps to the input image results in the final heatmap reflecting pixel-wise explanations. However, for point clouds, scaling is infeasible since points in a point cloud are unordered and precise association of all points to feature maps is not known. Our proposed algorithm effectively scales up sparse partial heatmaps to the input point cloud's size through the described iterative process.



\section{Experiments}
\label{sec:experiments}


\paragraph{Datasets and Implementation Details}
We evaluate our proposed approach APE on four different point cloud processing networks. Two fixed networks, namely PointNet and DGCNN, and two variable networks, name PointNet++ and VoteNet, are used to demonstrate our method's strength. We use three publically available datasets: ShapeNet-Part, Toy Flange, and SUN RGB-D dataset \cite{sunrgb}. The first is the ShapeNet-Part dataset \cite{shapenet} used for point cloud object classification with 16 categories of everyday objects. The second is the Toy Flange dataset which comprises of 2 categories of mechanical flanges of either 4 holes (157 files) or 8 holes (280 files).  This dataset will be made public. Finally, the SUN RGB-D dataset comprises of complete scene point clouds. The first two datasets are used to train PointNet, PointNet++, and DGCNN for classification. Our approach then generates heatmaps on the test set during inference. The third dataset is used to generate explanations on a pre-trained VoteNet \cite{votenet}. We build on existing open-source implementations \cite{gitvote,gitdgcnn,gitpointnet,gitgrad}. 

We compare our approach with existing techniques, namely Gradients \cite{sanity,gradients} and Point cloud Saliency Maps (PcSN). Gradients is an early method for visualising networks in the images domain, whereby the gradient values of the loss function w.r.t to image pixels are computed and visualised. We adapt this approach to the point cloud processing networks. Point cloud Saliency Maps (PcSN) \cite{pcSaliencyMap} is the state-of-the-art approach for computing contributions of each point towards the network's decision. We demonstrate that our approach APE outperforms both of these by generating heatmaps which not only assign reasonable relevance scores to individual points but are also intuitive for an observer as it highlights semantic parts of shape. 


\subsection{Qualitative Experiments}



\begin{figure*}
\begin{center}
    \includegraphics[width=1.0\textwidth]{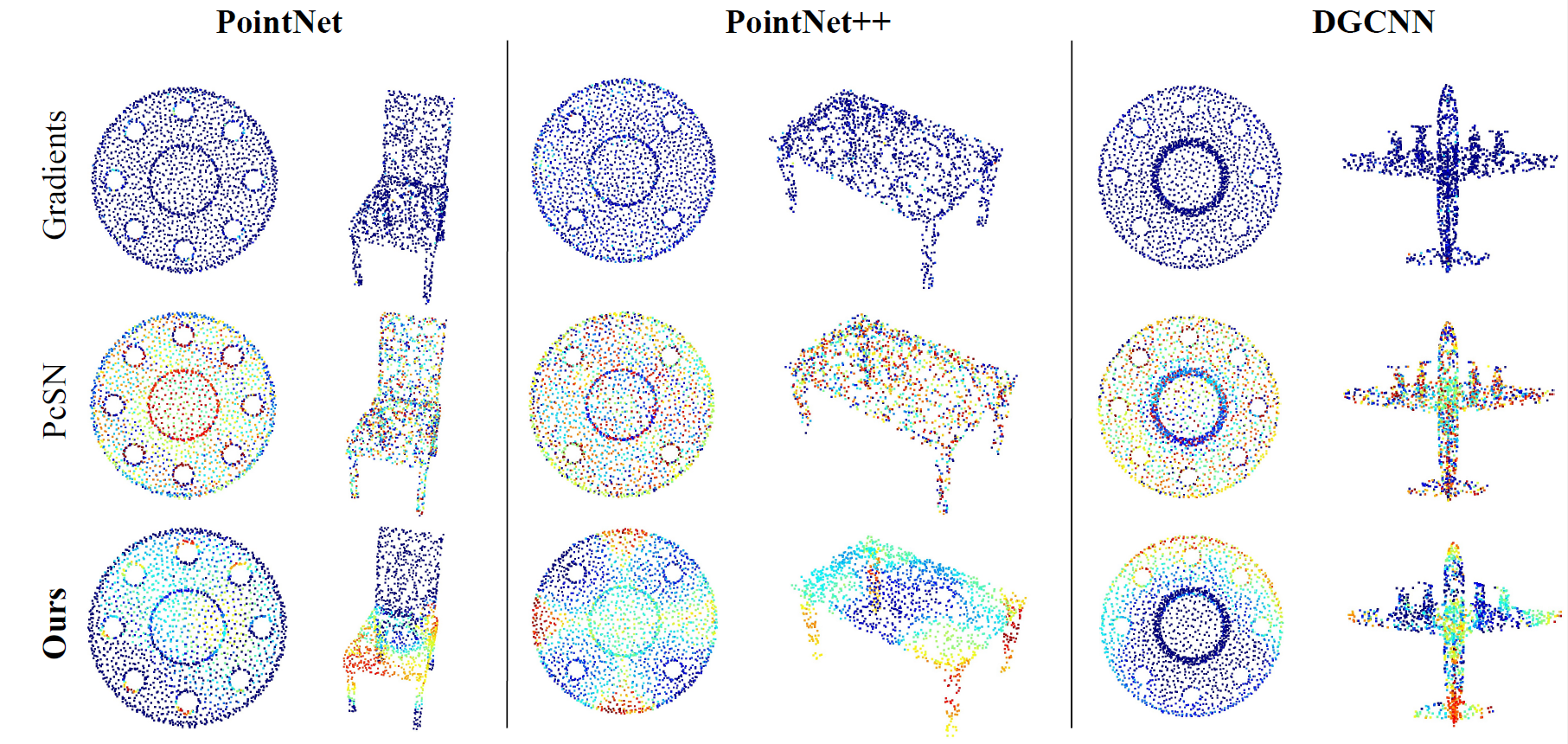}
\end{center}
 \vspace{-1em}
   \caption{\small{\small{{\bf Qualitative results of APE.} The visualization show randomly selected objects which have been correctly classified for PcSN \cite{pcSaliencyMap}, Gradients \cite{gradients} and \textbf{our} propose APE method. For each network, a Toy Flange dataset object is on the left, and a ShapeNet object is shown on the right.}}}
\label{fig:heatmap_overview}
\end{figure*}

\begin{figure*}
\begin{center}
   \includegraphics[width=1.00\textwidth]{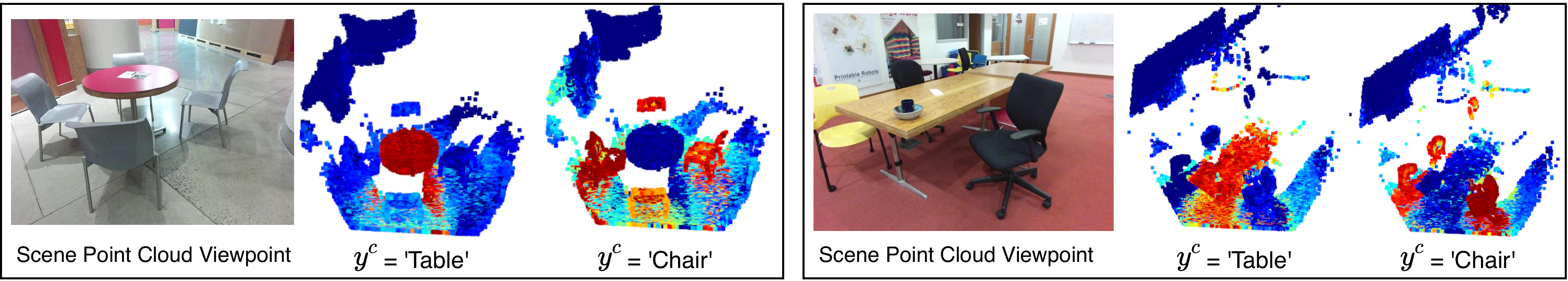}
\end{center}
\vspace{-1.0em}
   \caption{\small{{\bf Final Heatmaps for Scenes.} For VoteNet scenes, the APE approach results in point cloud heatmaps highlighting points belonging to the queried target class $y^c$. Note that highlighted points correspond to semantic categories of the objects of the selected target class.}}

\label{fig:votenetCAM}
\end{figure*}

\paragraph{Point Cloud Heatmaps Evaluation} Our APE approach was applied to different point cloud networks for object classification and detection. A sample of point cloud heatmaps generated from our approach as compared to other methods is presented in Fig. \ref{fig:heatmap_overview}. Point cloud heatmaps seen here assign a number to each point in the range [0,1], indicating low relevance (blue) to high relevance (red). Our method identifies significant segments of objects which are critical for decision-making for the network. It is apparent that our method generates highly interpretable heatmaps in comparison to Gradients and PcSN. Gradients approach produces a highly skewed explanation map where only a handful of points are identified as relevant whilst the majority of points remain insignificant. In contrast, PcSN creates an excessively high-resolution heatmap. Even though this method effectively identifies the most significant points, the resulting heatmaps are grainy and incomprehensible. Our method clearly generates intuitive point cloud heatmaps that can visually establish trust and faith in the networks.     

From our point cloud heatmaps, we notice that extremities or geometrically varied features, such as wingtips, table corners, table corners, hole's edges, etc. are clearly highlighted as significant, and planar surfaces, such as tabletops, floors, etc. are unremarkable. This is expected behaviour since planar surfaces lack geometric texture, which allows for distinguishing object classes apart. For VoteNet, see Fig. \ref{fig:votenetCAM}, it is shown that our approach is general for applicability over a large scene point cloud. In such large scenes, the method highlights points that correspond to semantic objects as set by the target class. We further note the target class objects within the scenes ($y^c=$`Chair') are spatially localized. Furthermore, our method's strength is apparent from the high precision of spatial boundaries between objects seen in VoteNet results. 
For brevity, multiple other example point cloud heatmaps for different objects have been presented in supplementary work. The results presented here show consistent superiority over the basic approach of Gradients \cite{gradients} and the state-of-the-art method PcSN \cite{pcSaliencyMap}.



\paragraph{Insight into Network's Decisions} Given the high veracity of point cloud heatmaps generated by our approach, it is possible to draw interesting insight into the network's decision-making process. Consider Fig. \ref{fig:heatmap_overview} DGCNN network architecture. Recall that the flange dataset poses a binary classification problem with 4-hole and 8-hole discrimination required. From visual inspection of our point cloud heatmaps in the figure, it is apparent that DGCNN and PointNet both have high focal points around the holes indicating correct network focus. A more interesting insight is that DGCNN heatmaps consistently show focus on only five out of eight holes for correct classification. This geometric feature is sensible for segregating between 4 and 8-hole binary class problem.
\begin{figure*}
\begin{center}
      \includegraphics[width=1.00\textwidth]{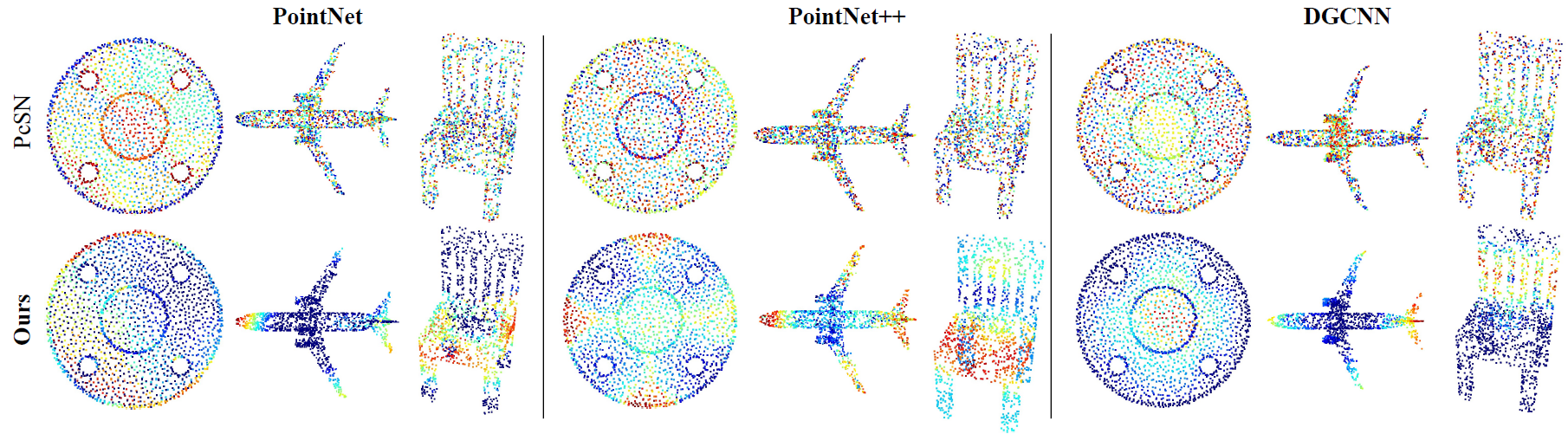}
\end{center}
   \vspace{-1em}
   \caption{\small{In-depth insight into various network architectures are revealed using our APE approach for generating heatmaps. For example, a clear incorrect focal point is seen for PointNet++ when classifying the 4-hole flange. Corresponding heatmaps from PcSN are shown to lack human interpretability.}}
\label{fig:heatmap_overview2}
\end{figure*}

We provide further examples of insights in Fig. \ref{fig:heatmap_overview2}. Most prominently, the PointNet++ architecture classifies the 4-hole flange by focusing on the empty spaces between the holes rather than the holes themselves. In other words, the network has learned to detect the absence of holes rather than their presence. This is a clear contradiction to common human reasoning when identifying a 4-hole flange. We also note that important sections of the chair from PointNet and PointNet++ are the seats, whereas DGCNN (having the most superior classification accuracy) has identified the unique pattern of the \textit{seat back} as a discriminative feature. This focal point allows DGCNN to discriminate better amongst other similar furniture in the dataset, e.g., sofas, thereby achieving higher accuracy. Finally, the shape airplane shows consistent focal points, e.g., nose, wingtips, and tail, across all network architectures. Such in-depth insights cannot be drawn from the PcSN heatmaps as they lack human interpretability evident from Fig. \ref{fig:heatmap_overview} and \ref{fig:heatmap_overview2}.

\begin{figure*}
\begin{center}
    \includegraphics[scale=0.11]{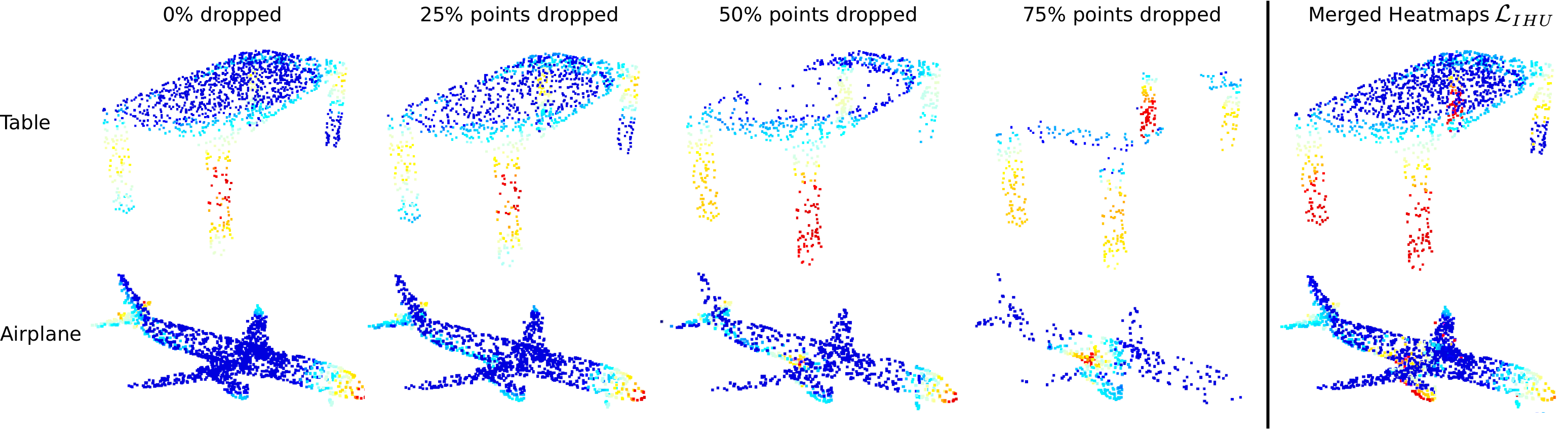}
\end{center}
   \vspace{-1em}
   \caption{\small{{\bf Iterative Heatmap Updating (IHU).} Visual example of the low-relevance dropping approach for PointNet at 0\%, 25\%, 50\%, and 75\% of points dropped. Note that the highest relevance points (red) differ as the heatmap is recalculated after each set of point drops.}}
\label{fig:visualpointdrop}
   \vspace{-1.5em}
\end{figure*}

\paragraph{IHU Low-relevance Point Dropping} Fig. \ref{fig:visualpointdrop} demonstrates the effect of dropping low-relevance points on heatmaps which are recalculated on the remaining points at every iteration. This is the outer loop of algorithm \ref{algo:APE}. Note that the highlighted regions change as points are dropped, indicating that a new set of points gain relevance for classification. It is further observed that more detailed explanations reveal on segments of objects in later iterations; for example, the table shows one leg being of high significance in the first column but all four legs are discovered as highly contributing segments with 75\% points dropped. The last column in Fig. \ref{fig:visualpointdrop} shows the merged point cloud, which incorporates all discovered explanations across the different iterations. These results confirm our assertion of dropping low-relevance points to recompute heatmaps iteratively and generate a more representative merged final heatmap.

\begin{figure*}
\begin{center}
      \includegraphics[width=1.0\textwidth]{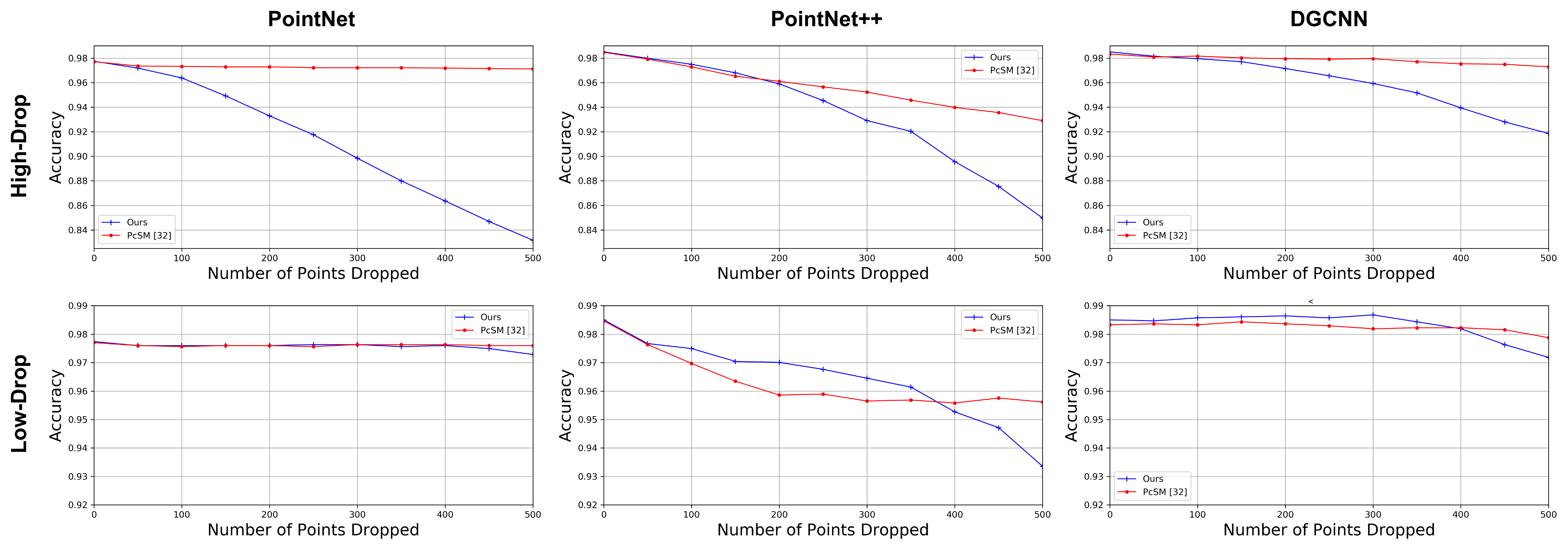}
\end{center}
   \vspace{-1.4em}
   \caption{\small{\textbf{Quantitative evaluation of APE.} The point dropping curve (PDC) has been calculated for various networks with our proposed and existing SOTA methods. The top row presents a high-drop experiment, whereas the lower row presents low-drop experiments. A consistent overall superiority over the existing method is evident across all networks.}}
\label{fig:QuantitativeGraph}
\vspace{-1.5em}
\end{figure*}

\subsection{Quantitative Experiments}

The point cloud heatmap obtained for various networks gives critical indications about the network's learned parameters. This point cloud heatmap can find critical points that can be used to assess our approaches quantitatively. In particular, for the classification task, the point cloud heatmap provides a way to determine the most and least relevant points in a given point cloud. To utilize the heatmap, we use the common evaluation measure point dropping curve (PDC) \cite{rise,pcSaliencyMap}. The PDCs show a drop in classification accuracy of the model as points are removed from the input point cloud. Points are dropped according to their computed relevance values, i.e., most relevant first (high-drop) and least relevant first (low-drop). The slope of these curves provides essential information about the quality of the heatmap generated. In particular, high-drop PDCs should fall steeper than the low-drop PDCs, and the accuracy should drop until its near-random guess. This metric can then be used to compare different approaches on the same network architecture.


\paragraph{Evaluation Results}

\begin{table}[ht]
\centering
\footnotesize

\begin{tabular}{|P{2.1cm}|P{1.8cm}|P{1.35cm}|P{1.85cm}|P{1.35cm}|}
 \hline
  Method & Drop Type &  PointNet &  PointNet++ &  DGCNN \\
 \hline
 Gradients \cite{gradients} & H.D.↓ & 0.90 & 0.70 & 0.93 \\
          & L.D. ↑ & 0.91 & \textbf{0.95} & 0.96 \\
 \hline
PcSN \cite{pcSaliencyMap} & H.D. ↓ & 0.89 & 0.80 & 0.93 \\
  & L.D ↑ & \textbf{0.92} & 0.83 & 0.96 \\
 \hline
 
 \textbf{Ours} & H.D. ↓ & \textbf{0.53} & \textbf{0.68} & \textbf{0.82} \\
 & L.D. ↑ & 0.91 & 0.85 & 0.96 \\
 
 \hline
\end{tabular}

\caption{\textbf{Area Under the Curve (AUC) } AUC has been calculated from the point dropping curves for all point dropping experiments. H.D. stands for high-drop and L.D. for low-drop. A higher value for L.D. and a lower value for H.D. is better.}
\label{tb:allresults}
\end{table}


 
 

Fig. \ref{fig:QuantitativeGraph} presents the PDCs for our method compared to the PcSN approach on both fixed-size (PointNet and DGCNN) and variable-size (PointNet++) networks. Note that we compute a final heatmap once for both appraoches and use these during the entire point dropping experimentation. Table. \ref{tb:allresults} present the area under the curve (AUC) values of the corresponding PDCs in Fig. \ref{fig:QuantitativeGraph}. It can be observed that when dropping high-contribution points (H.D.), the accuracy of the classification network significantly drops resulting in the AUC of our approach to be consistently lower than all comparative approaches on all networks. Furthermore, Fig. \ref{fig:QuantitativeGraph} illustrate this result whereby the high-drop curve of our approach sharply deteriorates compared to PcSN. This trend is indicative of the strength and correctness of our generated heatmaps as they have been assigned reasonable explanation values. In the case of PointNet++, our method's high drop line rises slightly above PcSN initially; however, the overall trend remains superior to PcSN. 

We also report results obtained from dropping low-contribution points (L.D.). We observe a mixed trend and neither method has superiority over the other using this experimental scheme as seen from Fig. \ref{fig:QuantitativeGraph} and table \ref{tb:allresults}. An exception is seen when comparing against the Gradients approach only on PointNet++, however, recall that qualitative results of Gradients (Fig. \ref{fig:heatmap_overview}) displays no \textit{meaningful} coloring of the point cloud heatmap. Human studies have been conducted to further validate the qualitative veracity of our approach, these are presented next.


\paragraph{Human Study:} We train a multi-label PN++ classifier with 6 object classes namely 'chair`, 'bike`,
'table`, 'plane`, 'car` and 'skateboard`. The input data for training and testing this classifier are not single object point clouds but rather two objects concatenated with a certain distance ensuring no overlap. For any given point cloud pair, we generate explanation maps, using different approaches, by setting the target class to one of the two objects in the input. For e.g. consider an input containing an airplane and a car, then given a target class as `airplane', generated heatmap is expected to highlight this class. We notice that our method consistently highlights the correct class in contrast to baselines. To confirm this we perform an extensive human study where each participant is asked to evaluate `which object demonstrate a better separation of object segments?' in a heatmap. Interestingly, our method comes out on top with the most correct classes corresponding with the user response as shown in Table. \ref{tb:ablation_study}. Further details on this study are included in the supplementary work.



\begin{table}[ht]
\centering
\footnotesize

\begin{tabular}{|P{1.4cm}|P{2.2cm}|P{1.8cm}|P{1.1cm}|}
 \hline
  &  Saliency Maps \cite{pcSaliencyMap} &  Gradients \cite{gradients} &  \textbf{Ours} \\
  \hline
  Accuracy & 0.44 & 0.51 & 0.72 \\
 \hline
\end{tabular}

\caption{\textbf{Human Study} Accuracies in this table display the percentage of human users who selected the correct response. These results indicate that our approach creates heatmaps which are interpretable by human subjects.}
\label{tb:ablation_study}
   \vspace{-1.5em}
\end{table}

\section{Conclusions}
\label{sec:conclusions}
In this work, we proposed a general approach to visually explain a wide variety of point cloud processing deep networks. We proposed the accumulated piecewise explanation (APE) algorithm, which tracks gradients to the final feature maps to generate a partial heatmap. This heatmap indicates the contribution of each point towards the network decision. Often, networks reduce and aggregate the features in subsequent layers. Partial heatmaps at later layers are mapped to the input point cloud size by iteratively computing explanations for segments of the input point cloud. These partial heatmaps are then concatenated to create a initial point cloud heatmap. This heatmap is then refined iteratively by dropping low-relevance points at each iteration to discover deeper explanations. We evaluate this approach against existing approaches and demonstrate good performance qualitatively and quantitatively. In the future, we aim to generalise to a broader range of network architectures and tackle networks designed for other types of unstructured data such as meshes or graphs.

\bibliographystyle{splncs}
\bibliography{egbib}

\begin{thebibliography}{10}

\bibitem{gradients}
Simonyan, K., Vedaldi, A., Zisserman, A.:
\newblock Deep inside convolutional networks: Visualising image classification
  models and saliency maps (2013)

\bibitem{CAM}
{Zhou}, B., {Khosla}, A., {Lapedriza}, A., {Oliva}, A., {Torralba}, A.:
\newblock Learning deep features for discriminative localization.
\newblock In: 2016 IEEE Conference on Computer Vision and Pattern Recognition
  (CVPR). (2016)  2921--2929

\bibitem{gradCAM}
Selvaraju, R.R., Das, A., Vedantam, R., Cogswell, M., Parikh, D., Batra, D.:
\newblock Grad-cam: Why did you say that? visual explanations from deep
  networks via gradient-based localization.
\newblock CoRR \textbf{abs/1610.02391} (2016)

\bibitem{lrp}
Binder, A., Montavon, G., Bach, S., M{\"{u}}ller, K., Samek, W.:
\newblock Layer-wise relevance propagation for neural networks with local
  renormalization layers.
\newblock CoRR \textbf{abs/1604.00825} (2016)

\bibitem{pointnet}
Qi, C.R., Su, H., Mo, K., Guibas, L.J.:
\newblock Pointnet: Deep learning on point sets for 3d classification and
  segmentation.
\newblock CoRR \textbf{abs/1612.00593} (2016)

\bibitem{pointnet++}
Qi, C.R., Yi, L., Su, H., Guibas, L.J.:
\newblock Pointnet++: Deep hierarchical feature learning on point sets in a
  metric space.
\newblock CoRR \textbf{abs/1706.02413} (2017)

\bibitem{dgcnn}
Wang, Y., Sun, Y., Liu, Z., Sarma, S.E., Bronstein, M.M., Solomon, J.M.:
\newblock Dynamic graph {CNN} for learning on point clouds.
\newblock CoRR \textbf{abs/1801.07829} (2018)

\bibitem{votenet}
Qi, C.R., Litany, O., He, K., Guibas, L.J.:
\newblock Deep hough voting for 3d object detection in point clouds.
\newblock CoRR \textbf{abs/1904.09664} (2019)

\bibitem{zeiler2014visualizing}
Zeiler, M.D., Fergus, R.:
\newblock Visualizing and understanding convolutional networks.
\newblock In: European conference on computer vision, Springer (2014)  818--833

\bibitem{rlgan}
Sarmad, M., Lee, H.J., Kim, Y.M.:
\newblock Rl-gan-net: {A} reinforcement learning agent controlled {GAN} network
  for real-time point cloud shape completion.
\newblock CoRR \textbf{abs/1904.12304} (2019)

\bibitem{panos}
Achlioptas, P., Diamanti, O., Mitliagkas, I., Guibas, L.J.:
\newblock Representation learning and adversarial generation of 3d point
  clouds.
\newblock CoRR \textbf{abs/1707.02392} (2017)

\bibitem{splat}
Su, H., Jampani, V., Sun, D., Maji, S., Kalogerakis, E., Yang, M., Kautz, J.:
\newblock Splatnet: Sparse lattice networks for point cloud processing.
\newblock CoRR \textbf{abs/1802.08275} (2018)

\bibitem{yolo}
Simon, M., Milz, S., Amende, K., Gross, H.:
\newblock Complex-yolo: Real-time 3d object detection on point clouds.
\newblock CoRR \textbf{abs/1803.06199} (2018)

\bibitem{imagenetData}
{Deng}, J., {Dong}, W., {Socher}, R., {Li}, L., {Kai Li}, {Li Fei-Fei}:
\newblock Imagenet: A large-scale hierarchical image database.
\newblock In: 2009 IEEE Conference on Computer Vision and Pattern Recognition.
  (2009)  248--255

\bibitem{coco}
Lin, T., Maire, M., Belongie, S.J., Bourdev, L.D., Girshick, R.B., Hays, J.,
  Perona, P., Ramanan, D., Doll{\'{a}}r, P., Zitnick, C.L.:
\newblock Microsoft {COCO:} common objects in context.
\newblock CoRR \textbf{abs/1405.0312} (2014)

\bibitem{modelnet}
Wu, Z., Song, S., Khosla, A., Tang, X., Xiao, J.:
\newblock 3d shapenets for 2.5d object recognition and next-best-view
  prediction.
\newblock CoRR \textbf{abs/1406.5670} (2014)

\bibitem{resnet}
He, K., Zhang, X., Ren, S., Sun, J.:
\newblock Deep residual learning for image recognition.
\newblock CoRR \textbf{abs/1512.03385} (2015)

\bibitem{imagenetCls}
Krizhevsky, A., Sutskever, I., Hinton, G.E.:
\newblock Imagenet classification with deep convolutional neural networks.
\newblock In Pereira, F., Burges, C.J.C., Bottou, L., Weinberger, K.Q., eds.:
  Advances in Neural Information Processing Systems 25.
\newblock Curran Associates, Inc. (2012)  1097--1105

\bibitem{cnn2}
LeCun, Y., Boser, B.E., Denker, J.S., Henderson, D., Howard, R.E., Hubbard,
  W.E., Jackel, L.D.:
\newblock Handwritten digit recognition with a back-propagation network.
\newblock In Touretzky, D.S., ed.: Advances in Neural Information Processing
  Systems 2.
\newblock Morgan-Kaufmann (1990)  396--404

\bibitem{objectlocfree}
{Oquab}, M., {Bottou}, L., {Laptev}, I., {Sivic}, J.:
\newblock Is object localization for free? - weakly-supervised learning with
  convolutional neural networks.
\newblock In: 2015 IEEE Conference on Computer Vision and Pattern Recognition
  (CVPR). (2015)  685--694

\bibitem{ziwen2019visualizing}
Ziwen, C., Wu, W., Qi, Z., Fuxin, L.:
\newblock Visualizing point cloud classifiers by curvature smoothing (2019)

\bibitem{pinherio}
Pinheiro, P.H.O., Collobert, R.:
\newblock Weakly supervised semantic segmentation with convolutional networks.
\newblock CoRR \textbf{abs/1411.6228} (2014)

\bibitem{rise}
Petsiuk, V., Das, A., Saenko, K.:
\newblock {RISE:} randomized input sampling for explanation of black-box
  models.
\newblock CoRR \textbf{abs/1806.07421} (2018)

\bibitem{graphGrad}
Pope, P.E., Kolouri, S., Rostami, M., Martin, C.E., Hoffmann, H.:
\newblock Explainability methods for graph convolutional neural networks.
\newblock In: The IEEE Conference on Computer Vision and Pattern Recognition
  (CVPR). (2019)

\bibitem{explain_pointnet}
Zhang, B., Huang, S., Shen, W., Wei, Z.:
\newblock Explaining the pointnet: What has been learned inside the pointnet?
\newblock In: The IEEE Conference on Computer Vision and Pattern Recognition
  (CVPR) Workshops. (2019)

\bibitem{qiu2019geometric}
Qiu, S., Anwar, S., Barnes, N.:
\newblock Geometric feedback network for point cloud classification (2019)

\bibitem{pcSaliencyMap}
Zheng, T., Chen, C., Yuan, J., Li, B., Ren, K.:
\newblock Pointcloud saliency maps.
\newblock In: Proceedings of the IEEE/CVF International Conference on Computer
  Vision. (2019)  1598--1606

\bibitem{sunrgb}
Song, S., Lichtenberg, S.P., Xiao, J.:
\newblock Sun rgb-d: A rgb-d scene understanding benchmark suite.
\newblock In: The IEEE Conference on Computer Vision and Pattern Recognition
  (CVPR). (2015)

\bibitem{shapenet}
Yi, L., Kim, V.G., Ceylan, D., Shen, I.C., Yan, M., Su, H., Lu, C., Huang, Q.,
  Sheffer, A., Guibas, L.:
\newblock A scalable active framework for region annotation in 3d shape
  collections.
\newblock SIGGRAPH Asia (2016)

\bibitem{gitvote}
facebookresearch:
\newblock Deep hough voting for 3d object detection in point clouds.
\newblock \url{https://github.com/facebookresearch/votenet} (2020)

\bibitem{gitdgcnn}
Wang, Y.:
\newblock Dynamic graph cnn for learning on point clouds.
\newblock \url{https://github.com/WangYueFt/dgcnn} (2019)

\bibitem{gitpointnet}
Xia, F.:
\newblock Pointnet.pytorch.
\newblock \url{https://github.com/fxia22/pointnet.pytorch} (2019)

\bibitem{gitgrad}
Jiaxin, L.:
\newblock Grad-cam implementation in pytorch.
\newblock \url{https://github.com/jacobgil/pytorch-grad-cam} (2020)

\bibitem{sanity}
Adebayo, J., Gilmer, J., Muelly, M., Goodfellow, I.J., Hardt, M., Kim, B.:
\newblock Sanity checks for saliency maps.
\newblock CoRR \textbf{abs/1810.03292} (2018)

\end{thebibliography}

\end{document}